\documentclass[10pt,a4]{article}
\usepackage{amsmath,amssymb}
\usepackage{url}
\usepackage{graphicx}
\usepackage{subfigure}
\usepackage{cite}
\usepackage{color}
\usepackage{comment}
\usepackage[toc,page]{appendix}

\begin{document}

\title{Causality, Information and Biological Computation: An algorithmic software approach to life, disease and the immune system\footnote{Invited chapter contribution to \textit{Information and Causality: From Matter to Life}. Sara I. Walker, Paul C.W. Davies and George Ellis (eds.), Cambridge University Press. Some sections of this chapter were based on an invited keynote lecture H.Z. delivered in Vienna, Austria at the Complexity Science Hub conference \textit{Understanding Complexity, Offering Solutions to Problems of the 21st Century} in February 2015.
}}

\author{Hector Zenil\footnote{hector [dot] zenil [at] algorithmicnaturelab [dot] org}, Angelika Schmidt and Jesper Tegn\'er\\
Unit of Computational Medicine, Department of Medicine Solna,\\
Centre for Molecular Medicine, Science for Life Laboratory\\
(SciLifeLab), Karolinska Institute, Stockholm, Sweden}
\date{}

\maketitle

\begin{abstract}

Biology has taken strong steps towards becoming a computer science aiming at reprogramming nature after the realisation that nature herself has reprogrammed organisms by harnessing the power of natural selection and the digital prescriptive nature of replicating DNA. Here we further unpack ideas related to computability, algorithmic information theory and software engineering, in the context of the extent to which biology can be (re)programmed, and with how we may go about doing so in a more systematic way with all the tools and concepts offered by theoretical computer science in a translation exercise from computing to molecular biology and back. These concepts provide a means to a hierarchical organization thereby blurring previously clear-cut lines between concepts like matter and life, or between tumour types that are otherwise taken as different and may not have however a different cause. This does not diminish the properties of life or make its components and functions less interesting. On the contrary, this approach makes for a more encompassing and integrated view of nature, one that subsumes observer and observed within the same system, and can generate new perspectives and tools with which to view complex diseases like cancer, approaching them afresh from a software-engineering viewpoint that casts evolution in the role of programmer, cells as computing machines, DNA and genes as instructions and computer programs, viruses as hacking devices, the immune system as a software debugging tool, and diseases as an information-theoretic battlefield where all these forces deploy. We show how information theory and algorithmic programming may explain fundamental mechanisms of life and death.

\end{abstract}

\section{Introduction}

Information and computation have transformed the way we look at the world beyond statistical correlations, the way we can perform experiments, through simulations, and the way we can test these hypotheses. In his seminal paper on the question of machine intelligence~\cite{turingai}, Turing's approach consisted in taking a computer to be a black box and evaluating it by way of what one could say about its apparent behaviour. The approach can be seen as a digital version of a \emph{cogito, ergo sum} dictum, acknowledging that one can only be certain of one's own intelligence but not of the intelligence of anyone or anything else, including machines. This may indeed amount to a kind of solipsism, but in practice it suggests tools that can be generalised. This is how in reality we approach most areas of science and technology, not just out of pragmatic considerations but for a fundamental reason. Turing, like G\"{o}del  before him~\cite{godel}, showed that systems such as arithmetic that have a certain minimal mathematical power to express something, can produce outputs of an infinitely complicated nature~\cite{turing}. Indeed, this means that while one can design software, only trivial programs can be fully understood analytically to the point where one is able to make certain predictions about them. In other words, only by running software can one verify certain of its computational properties, such as whether or not it will crash. These fundamental results deriving from both G\"{o}del's incompleteness theorem and Turing's halting problem point in the direction that Turing himself took in his ``imitation game", which  we now identify as the  \emph{Turing test}.

Turing's halting problem implies that one cannot in general prove that a machine will ever halt, or that a certain configuration will be reached upon halting; one has therefore to proceed by testing and only by testing. These fundamental theorems and results imply that testing is unavoidable; even under optimal conditions there are fundamental limits to the knowledge one can obtain simply by looking at a system, or even by examining its source code, that is, its ultimate causes. Even with a knowledge of all the causes driving a system's evolution, there are strong limits to the kind of understanding of it that we can attain.

In light of these realities, entire (relatively) new fields designated \emph{model checking}, \emph{systems testing} and \emph{software verification} seek to produce and test reliable software based on simple mathematical models. Today's airplane construction companies and other manufacturers of critical systems such as electronic voting systems use these tools all the time; they are a minimal operational requirement in these industries. Biological phenomena are likely to be at least as complicated, and our understanding of them subject to the same knowledge limitations. That is, the proof that nature and most of our models of nature can contain systems such as arithmetic and digital computers constitutes a limit to our knowledge. One must then acknowledge and deal with the fact that there are true and unavoidable limits to knowledge extraction from artificial but also from natural systems, even under ideal circumstances.

\begin{figure}
\centering
\scalebox{.33}{\includegraphics{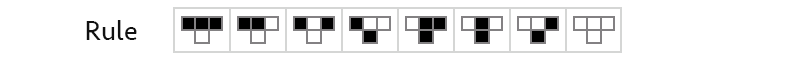}}\\

\bigskip

\scalebox{.2}{\includegraphics{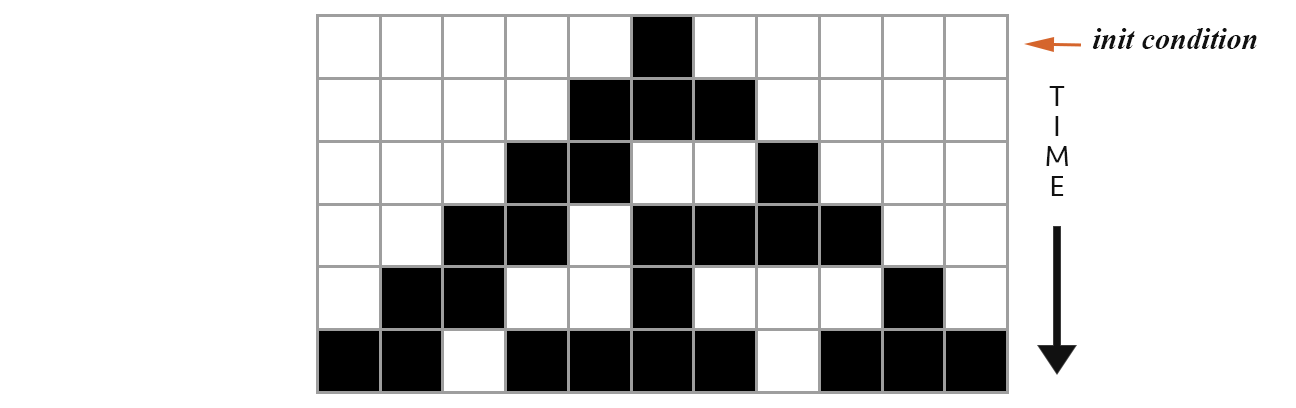}}
\caption{\label{fig1}There is a common misconception that complexity is about size, number of interacting parts or agents, and number of interactions. In this example, a minimalistic computer program starts from the simplest possible situation, yet produces random-looking (statistical) behaviour that is for all practical purposes unpredictable. This computer program, called a \textit{cellular automaton} (or CA), has a rule with code 30 (the rule icon represented in binary) according to Wolfram's enumeration~\cite{nks}. The first row represents the input of the program. Every black or white cell is updated according to the rule icon on top, which according to the 2 nearest neighbours of the cell, proceeds to change or retain the colour of the central cell in the next row.}
\end{figure}

\begin{figure}
\centering
\scalebox{.34}{\includegraphics{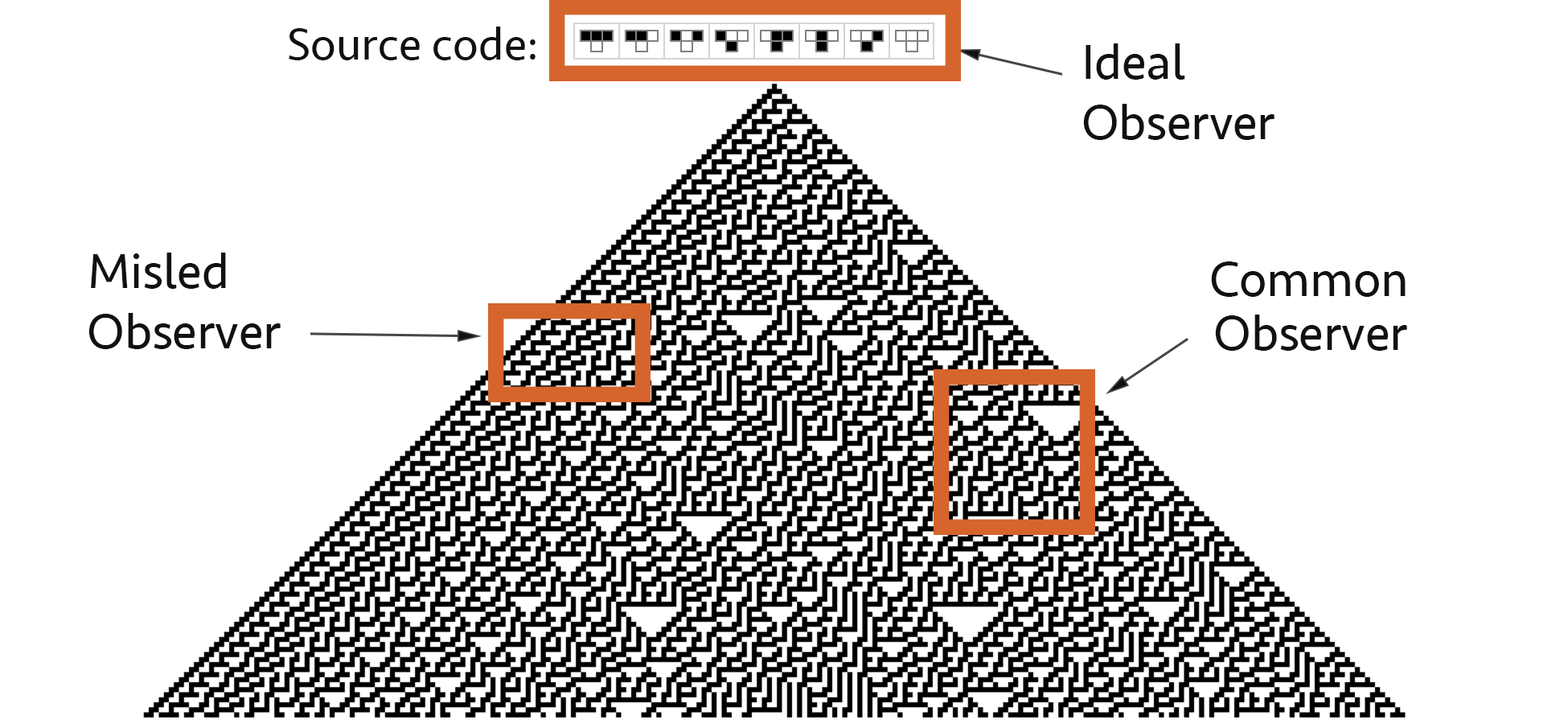}}

\caption{\label{fig2}A simple computer program such as elementary CA rule 30, with the simplest possible initial condition, can give us a sense of how complex we may expect natural phenomena to be, and it helps illuminate scientific practice and its limitations. Despite deriving from a very simple program, without knowing the \textit{source code} (top) of the program, a partial view and a limited number of observations can be misleading or uninformative. For example, the misguided observer may think that the system is very simple and that the source code can only generate regular behaviour, but would in fact be missing most of the picture. This example shows how difficult it can be to reverse engineer a system from its evolution in order to discover its generating code when one is only privy to partial and limited sections of the system's evolution.}
\end{figure}

As shown in Figs.~\ref{fig1} and~\ref{fig2}, for an extremely simple computer program represented by an elementary cellular automaton (ECA), one needs to perform, in the best case scenario, at least 8 very precise observations at two consecutive times (hence 16) with perfect accuracy in order to hack this computer program, only to unveil its ridiculously small source code (which can be written in a few bits) and determine that the producing rule is ECA rule 30. Missing just a single observation can lead one to a rule with completely different behaviour, as they are highly sensitive, and by making more observations one immediately starts overfitting, leading one to a more complicated rule that may produce the same behaviour but is ultimately different.

Fig.~\ref{fig1} shows such a minimalistic example that it suggests that the same phenomenon is pervasive in physics and biology, where even the simplest conditions can generate a cascade of apparent randomness. Notice this is of even more basic nature than the phenomenon of \emph{chaos} where the argument is that arbitrary close initial conditions diverge in behaviour over time, an additional complexity in, for example, experimental reproducibility.

Multiple sclerosis (MS), for example, is a complex disease in which the insulating shield (the myelin sheath) of nerve cells is damaged and for which there is currently no curative treatment, yet the disease sometimes shows apparent periodic behaviour in the form of relapses. Relapsing eventually becomes progressive meaning that there is no further apparent partial recovery of the patient over time. In some cases, this might be a conflation of a misled observation coming from a compensation mechanism of the brain to take control over damaged neuromotor function rather than a strengthening of the immune system or a partial restoration of the neuron's myelin. The observation may be leading to the wrong conclusion that there are two types of MS when there may be, perhaps, only one behaving in different ways in different times.

\begin{figure}
\centering
\scalebox{.26}{\includegraphics{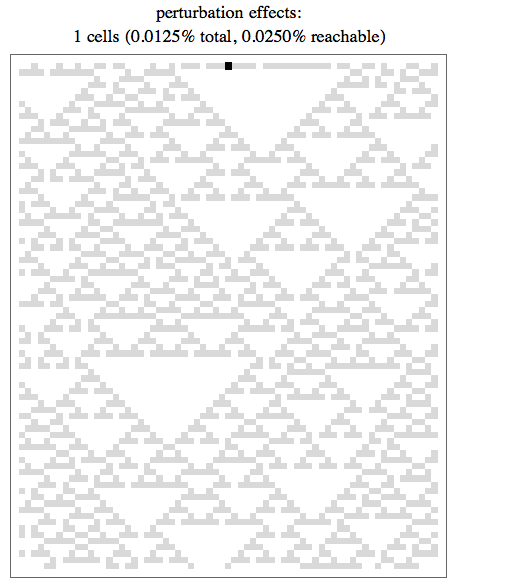}} \scalebox{.26}{\includegraphics{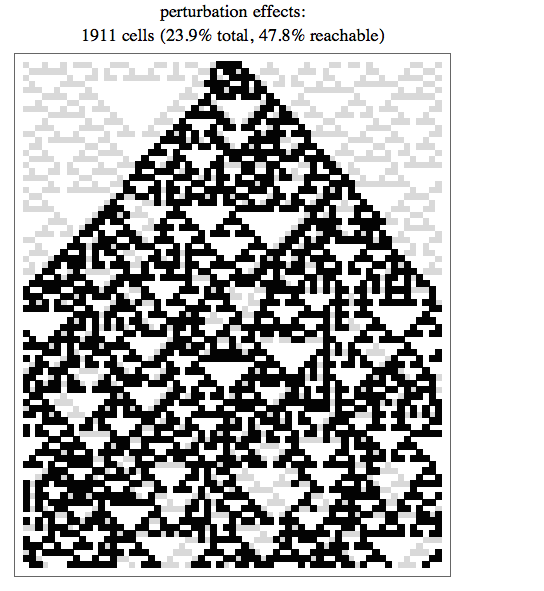}}\\
\bigskip
\bigskip

\scalebox{.28}{\includegraphics{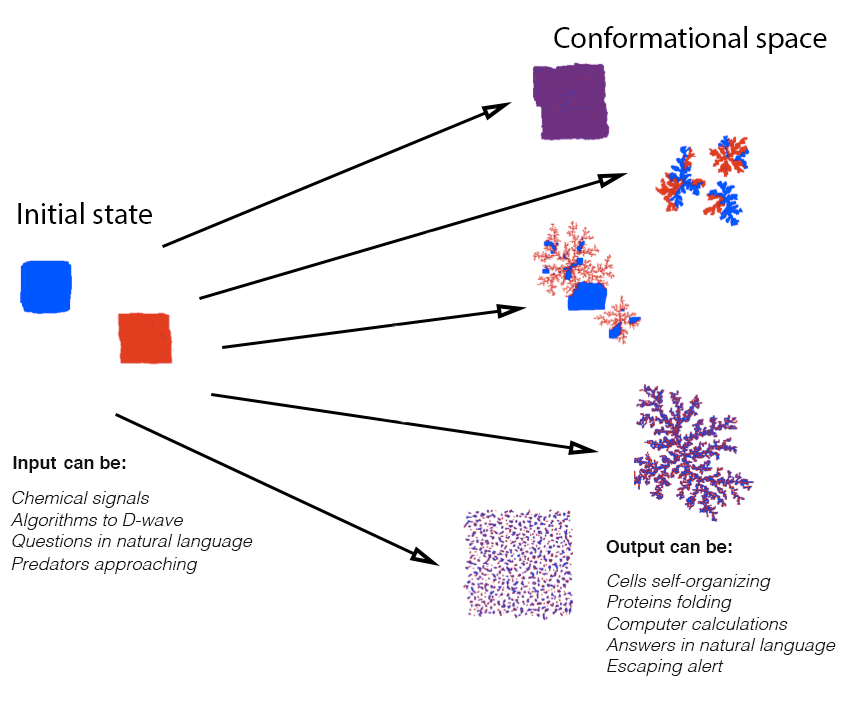}}
\caption{Top: Behaviour shown by ECA rule 22 for 2 different small perturbations. Different perturbations have different effects on the same computer program. They can either have no impact or can propagate at maximum speed, making the system appear highly sensitive and  chaotic. Bottom: Dealing with a black box. In this example of what can happen in and with an artificial or natural system in respect of which it is practically and fundamentally impossible to analytically determine all possible outputs for all possible inputs, we used~\cite{terrazas} a Wang tile-based computing model to map a simplified parameter space of emulated physical properties determining the binding capabilities of two (assumed) types of biological molecules simulating the behaviour of porphyrin molecules to the conformation space generated when the molecules interact.}
\label{conformationalspace}
\end{figure}

An ideal observer would be able to see the source code directly, but this is virtually impossible in practice, for a number of reasons. First, because it is difficult to separate phenomena from other phenomena in nature, and second because we never have access to first causes, or for that matter, to first causes in complete isolation (except perhaps in very artificial and controlled cases, such as are almost never encountered in nature). Moreover, true or apparent noise (e.g. measurement limitations) can have dramatic effects (see Fig.~\ref{conformationalspace} bottom). The scientific method deals with these fundamental problems, but it does so for even more fundamental reasons than is usually believed, as these problems actually characterise even the most minimalistic and fully deterministic systems (such as these computer programs used for purposes of illustration). In other words, it is practically impossible to recover source codes in nature under noisy, imperfect and limited conditions. And even if we manage to do so, that does not mean we can fully understand the behaviour of a system from its source code, just as nobody could imagine the complexity that rule 30 would be able to produce and nobody has been able to understand its evolution and find significant computational shortcuts. 

This phenomenon is not exclusive to rule 30 but is pervasive even in the most deterministic algorithmic sciences such as mathematics and computer science, in objects such as the decimal expansion of numbers like $\pi$ or the square root of 2, or in the logistic map that leads to chaos behaviour. But notice that these chaotic systems are of a slightly different nature. They are complex only when taken together with the description of their initial conditions. Here, rule 30 or the digits of $\pi$ look statistically random for all intents and purposes, even though these are objects for which there is no complex initial condition or else they start from the simplest initial condition, unlike initial conditions encoded in real numbers and in descriptions of ``close initial conditions". In nature there are no \textit{ideal observers} like the possibly lucky one circled in Fig.~\ref{fig2}. 

In other words, as Fig.~\ref{fig2} illustrates, reverse engineering (finding ultimate rules) is extremely difficult, and in biology there is an entire relatively new field devoted to reverse engineering biological networks, also called network reconstruction, where rules involve genes, proteins and metabolites, to name a few instances. If a system such as rule 30, despite its incredible simplicity, can be relatively difficult to reverse-engineer, one can only speculate as to the complication of reverse-engineering real-world systems such as (indirect)  gene-gene interactions (through gene products) from gene expression maps, that is, ascertaining which genes are (inter-)connected/coregulated with which others, down- or up-regulating their expressions and determining which proteins to produce for different cellular functions.

\section{Natural computation and programmability}

Sensitivity analysis can be useful for testing the robustness and variability of a system, and it sheds light on the  relationship between input and output in a system. This is not hard to identify from a computing perspective, where inputs to computer programs produce an output, and computer programs can behave in different ways. 

\begin{figure}
\centering
\scalebox{.3}{\includegraphics{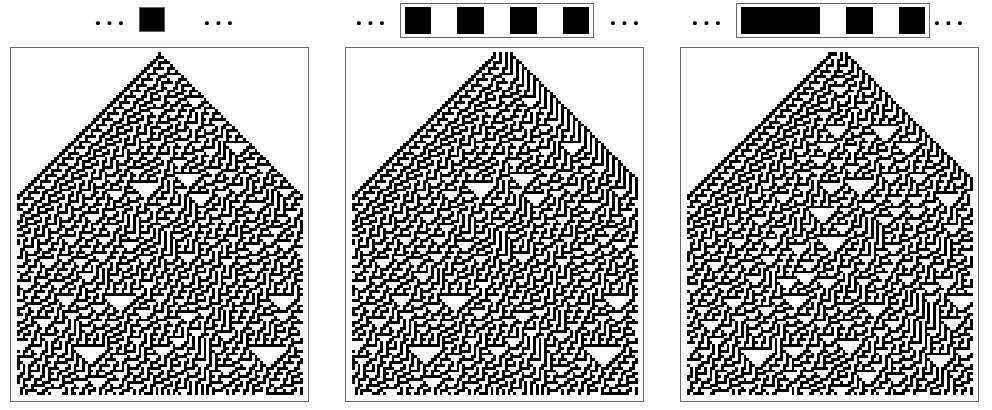}}

\bigskip

\scalebox{.3}{\includegraphics{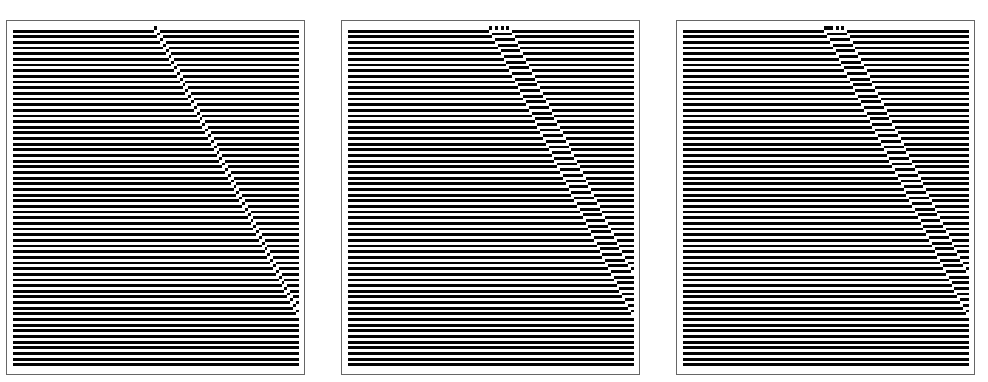}}

\bigskip

\scalebox{.3}{\includegraphics{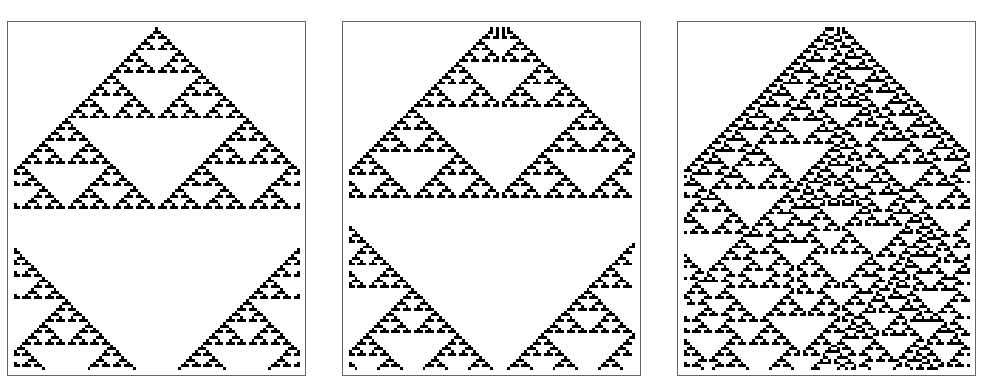}}

\caption{\label{fig3}Rule 30 (top), rule 3 (middle) and rule 22 (bottom) ECAs showing that with slightly different initial conditions two of them lead to the same qualitative behaviour and therefore have low \textit{programability}, whereas rule 22 shows greater variability and sensitivity to small perturbations and can therefore be better controlled to perform tasks than can rules 3 and 30. This (in)sensitivity can be explored with an online program available at~\cite{zenildemo}.}
\end{figure}

For example, sensitivity measures aim to quantify this uncertainty and its propagation through a system. Among common ways to quantify and study this phenomenon is, for example, the so-called Lyapunov exponent approach~\cite{lyapunov}. This approach consists in looking at the differences that arbitrarily close initial conditions produce in the output of a system, and hence is similar to the generalisation we introduce. Traditionally, if the exponent is large the sensitivity is non-linear, and divergence increases over time. If constant, however, the system is simple under this view. Programming systems requires them to have a non-zero Lyapunov exponent value, but in general this measure is not quite suitable, as it quantifies the separation of infinitesimally close trajectories which may be ill-defined in discrete systems, and traditionally, inputs for a program are not intended to be \textit{infinitesimally close}, as a distance measure would then be required (even though in~\cite{zenilca} one has been proposed). Furthermore, measures such as the Lyapunov exponent are meant to detect qualitative changes such as chaotic behaviour, which is not desirable for a programmable system, as is suggested by Fig.~\ref{fig3}. Further research on the connections between programmability and these other dynamical system sensitivity measures needs to be undertaken in the future. But they only seem adequate for quantifying far greater quantitative changes than are needed for a system to be programmable while at the same time not being too sensitive to small qualitative changes (e.g. statistical, such as systems that look random even if the rate of cell changes is maximal, as in the example shown in Figs.~\ref{fig1} and~\ref{fig2}(top)). Advances in natural and unconventional computation have made it difficult to measure how a computer or system behaves, and it is necessary to develop new tools for the purpose. In recent years there has been a heated debate concerning a company called D-wave, which claims to have built a quantum computer that it is selling as such, based on a technique called quantum annealing. But as we have seen, evaluating whether something actually is what it was designed to be is often difficult, and interestingly, this is another case in which the said quantum computer has been evaluated as a black box~\cite{dwave}.

In our approach to behaviourally evaluate systems for their programmability, a compression algorithm can be used as an interrogator device, where for questions one uses initial conditions of the system, while the answers are the lengths of the compressed outputs. In general, for a system to be reprogrammable, inputs with different information content should lead to outputs with different information content.  If the system reacts to these stimuli in a non-trivial fashion and passes the compression/complexity test, then one should be able to declare it capable of computing and amenable to being programmed; otherwise it is merely a system which produces similar answers (outputs) irrespective of the questions (inputs). To test this variability one compares the information content of the answers. This means the answers can look very different but if they contain the same information  they will fail the test. A system that does not react to the stimuli and produces trivial output will also clearly fail the test, and a system that only produces random output will fail too, because the information content of the answers, given their differences, will cancel each other out.This is then related to another concept in formal software-engineering, the concept of \emph{code coverage}, a measure that describes the degree to which the source code of a program is tested by a particular test suite. A program with high code coverage has been more thoroughly tested and has a lower chance of containing software bugs than a program with low code coverage. However, code coverage, and all other software measures traditionally require us to know or have access to the original code.  We are setting forth ideas for approaching situations when this code is not only difficult, but in principle impossible to know, because there is no way to rule out alternative codes that may be concurrently performing other computations for which no tests have been designed.

These testing ideas are based on whether a system whose source code may never be known is capable of reacting to the environment --- the input --- as is the case in more formal implementations of this measure of \textit{programmability}. Such a measure~\cite{zenilpt,zeniljetai} would quantify the sensitivity of a system to external stimuli and could be used to define the amenability of a system to being (efficiently) programmed. The basic idea is to replace  the observer in Turing's imitation game with a lossless compression algorithm, which duplicates the relevant subjective qualities of a regular observer, in that it can only partially ``detect" regularities in the data which the algorithm determines, there being no effective (programmable in practice) universal compression algorithm, as proven by the uncomputability of Kolmogorov complexity~\cite{kolmo,chaitin}.
The compression  algorithm looks at the evolution of a system and determines, by means of feeding the system with different initial conditions (which is analogous to questioning it), whether it reacts to external stimuli. The Kolmogorov complexity of an object is the length of the shortest program that outputs the object running on a universal computer program (that is, a computer program that can run any other computer program; the existence of such programs having been proved by Turing himself~\cite{turing}). Then, if the evolution of a program (say a natural phenomenon) is complex and does not react to external stimuli, all compression algorithms will fail at compressing its evolution and no difference between different evolutions for different perturbations will be detected (e.g. by taking the differences between the compressed lengths of the objects for the 2 perturbations, see Fig.~\ref{fig3}).

For example, as is shown in~\cite{zenilca}, certain elementary cellular automata rules that are highly sensitive to initial conditions, and present phase transitions which dramatically change their qualitative behaviour when starting from different initial configurations, can be characterised by these qualitative properties. A further investigation of the relation between this transition coefficient and the computational capabilities of certain known (Turing) universal machines has been undertaken in \cite{zenilca}. Other calculations have been advanced in~\cite{zenilpt} and ~\cite{zeniljetai}.

\begin{figure}
\centering
 \scalebox{.212}{\includegraphics{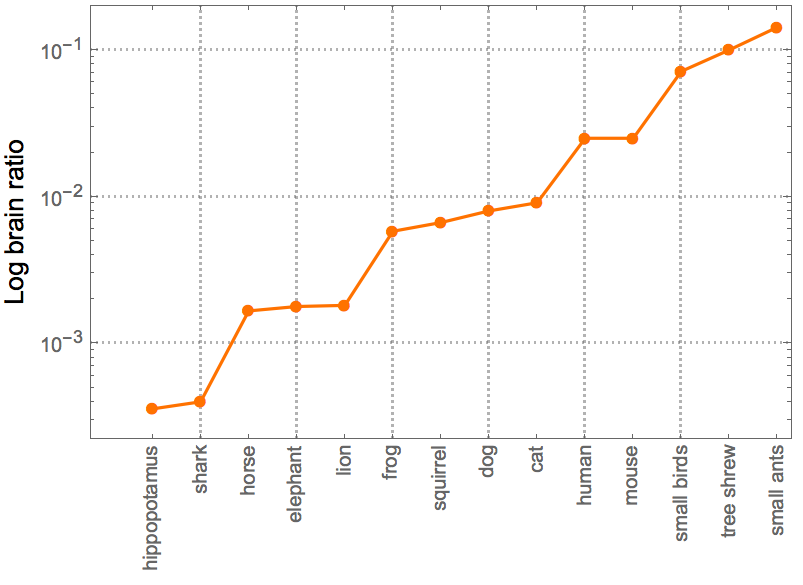}} \scalebox{.21}{\includegraphics{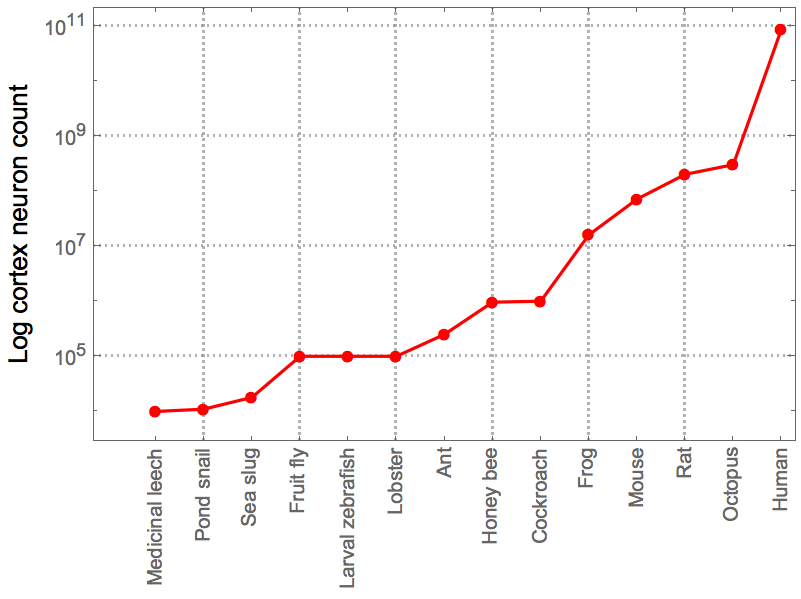}}\\
 \bigskip
 \bigskip
 
\scalebox{.26}{\includegraphics{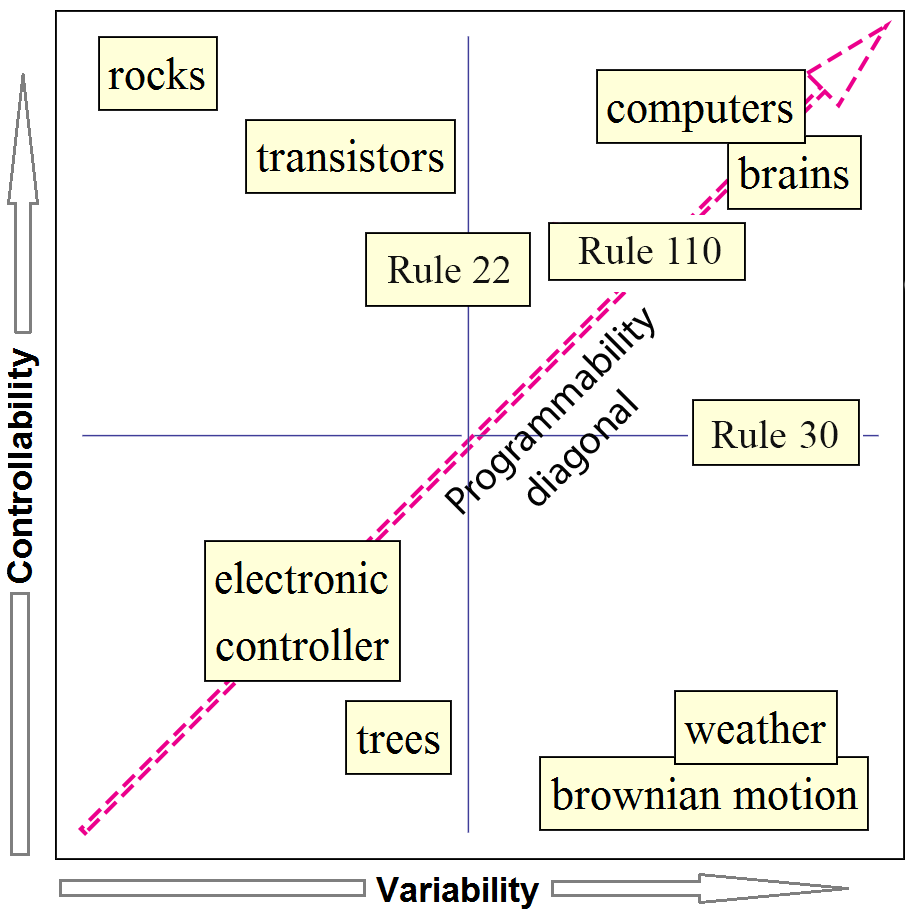}}
\caption{Top: Information is key to living organisms and therefore information theory is essential to understanding them. For example, while neither cranial capacity nor brain-to-body ratio (upper left) correspond to any biological feature, intelligence is clearly seen to be a function of information processing, when sorting species by cerebral cortex neurons count (upper right). Compiled data sources:~\cite{wiki1} and~\cite{wiki2}. Bottom: Programmability is a combination of variability and controllability~\cite{zenilball}. Behavioural measures suggest a natural classification of the programmability of natural and artificial systems, just as they suggest a classification of tumours (see Section~\ref{classificationcancer}) by degree of programmability based on their sensitivity to their environment and external stimuli and their variability. This reclassification deletes some artificial boundaries between living and non-living systems when it comes to behavioural and computing capacities, and between tumour types that are classified by tissue of origin rather than in terms of the error in the cell replication process that produces them.}\label{programmability}
\end{figure}

The behavioural approach in fact generates a natural classification of objects in terms of their programmability, as sketched in Fig.~\ref{programmability} (bottom). For example, while weather phenomena and Brownian motion have great variability, they are hardly controllable. On the other hand, rocks have a very low variability and hence are trivially controllable but are not therefore on the programmability diagonal and cannot count as computers. Everything on the diagonal, however, including living organisms, is programmable to some extent.

\section{Information biology}

One of the aims of our research is to exploit these ideas in order to try to reprogram living systems so as to make them do things we would like them to do, for in the end this is the whole idea behind programming something. Fig.~\ref{conformationalspace} (bottom) is an illustration of an investigation~\cite{terrazas} into the possibility of mapping the conformational space of a simulation of porphyrin molecules with a view to making them self-organise in different ways. That is, an investigation into what it means to program a nature-like simulated system, to find the inputs for the desired output (that matched the actual behaviour of the molecule). The challenge is about mapping the parameter space to the output space, in this case the conformational space of porphyrin molecules, i.e. the space of its possible shapes under certain conditions.

\begin{table}[htp]
\caption{\label{table1}Properties of simple \textit{versus} complex diseases. A good illustration of these differences can be found in~\cite{peltonen}. They cannot be considered in isolation. For example, the presence of many interacting elements does not automatically imply a complex disease (system); it is the way they interact and the multiple factors involved that make for a complex and therefore unpredictable system, even if it is fully deterministic (which may or may not be the case for complex diseases).}
\begin{center}
\begin{tabular}{c|c}
\hline
\textbf{Simple diseases} & \textbf{Complex diseases}\\
\hline
- Single and isolated causes & - Multiple disjoint and \\
- Simple gene expression patterns &joint causes\\
- Focalised effects & - Many interacting elements\\
- High predictability & - External factors\\
 & - Low predictability\\
\textbf{Examples:} & \\
Monogenic (single gene) diseases & \textbf{Examples:}\\
 (e.g. Cystic fibrosis), chromosomal & Multiple sclerosis, Alzheimer,\\
 abnormalities, Thalassaemia, Haemophilia,&Parkinson, most cancers\\
 Huntington's disease &\\
\hline
\end{tabular}
\end{center}
\label{default}
\end{table}%

 In biology, the greatest challenge is the prediction of behaviour and shape (``shape" determines function in biology). Examples are protein folding or predicting whether immune cells will differentiate in one direction rather than another. In Fig.~\ref{conformationalspace} (bottom), we investigated how we could arrive at certain specific conformational configurations from an initial state by changing environmental variables, such as temperature and other factors influencing binding properties. The proofs that certain abstract systems implemented in software can reach Turing universality constitute one example of how hardware may be seen as software. Indeed, by taking simple four-colored tiles (called Wang tiles, after Hao Wang) and placing them on a board according to the rule that whenever two tiles touch the sides that touch must be of the same colour, one can build an abstract machine that can simulate any other computer program. And this is a very powerful approach because one can program molecules or robots to do things like carry a payload to be released when certain specific conditions are met, release a chemical that may be used as a biological marker, fight disease or deal with nuclear waste. While we cannot try every possible perturbation, either on computer programs or on natural systems, and therefore require some fundamental analytical approach,  collections of molecular profiles from thousands of human samples, as made available by large consortia such as TCGA, TARGET, and ICG, among others, is making this process possible, because perturbations are a common mechanism of nature and natural selection.

This means that with enough data one does not need to perform perturbations but rather to find them as they occur in nature. However, while in recent decades we have come to understand that complex systems such as the immune system cannot be thoroughly analysed and fully understood as regards its components and function by using ``reductionist" approaches, and we have moved to the science of systems, such as systems- and computational biology, we have nevertheless failed to make as radical a move as we ought to and apply the tools of complexity science in a holistic way. And reprogramming cells to fight complex diseases (see Table~\ref{table1}) is one of our ultimate goals.

Complex diseases (see Table~\ref{table1}) require a complex debugging system, and the immune system can be viewed to play just this role. The immune system is the highly complex and dynamic counterpart of many complex (and simple) diseases, and the most common approach to understanding it has been \textit{via} evaluating its individual components. However, as with computer programs, this kind of approach cannot always expose the way in which an immune system works under a range of possible circumstances, particularly those leading to complex diseases such as autoimmune conditions and cancer. One instance where our knowledge will come up against limitations imposed by the systems approach itself is the process by which the immune system responds to infectious agents by activating innate inflammatory reactions and dictating adaptive immune responses responses (the immune system is split into two branches: innate and adaptive immune system). A similar example is how cancer systems biology has emerged to address the increasing challenge of cancer as a complex, multifactorial disease but has failed to move away once and for all from the archaic classification of cancer by tumour tissue of origin. 

Interestingly, these two highly complex systems are interconnected  ---  the immune system plays a key role in tumour development: The immune status (``immunoscore") was found to be the highly predictive for cancer patient survival~\cite{galon}, and evading immune destruction was added to the ``hallmarks of cancer" recently~\cite{hallmarks}. According to the immune surveillance theory, the immune system does not only recognise and combat invading pathogens but also detects host cells that become cancerous, thus eliminating tumor cells as soon as they arise. This requires the immune system to detect ``abnormal" structures on a tumour cell that is otherwise ``self"  ---  despite self tolerance as an important hallmark of the immune system. In fact, the immune system can be viewed as another example of the black-box behavioural approach to systems. Immune cells ``expect" certain responses from the cells they encounter in the body, using these to identify their nature. When they do not recognise these signals as harmless, they attack the cells. In fact, sometimes this process fails and they attack the wrong cells, the healthy cells of the host body that are not by any means foreign or abnormal; this defines what an autoimmune disease is. It is yet another example of a programmed function of the cell going wrong and requiring reprogramming by none other than external signals (which is the way any system, including a computer, is reprogrammed by giving them a program to read, a signal).

Equipped with these ideas that suggest that we can extend concepts that properly belong only to computability and algorithmic information theory, we can devise a software-engineering approach to systems biology.

\subsection{How natural selection programs and reprograms life}

At some point early in the process of replicating (copying) biological information for cell growth, which is particularly necessary for multicellular organisms, the process reaches the critical juncture of dealing with errors and redundancy that can be quantified by the so-called Noisy-channel coding theorem, part of Shannon's classical information theory, determining the degree of redundancy needed to be able to deal with different degrees of noise.

Had the balance not been reached, no copying process would have conveyed the information necessary for organisms to reproduce. The noisy-channel coding theorem (sometimes called Shannon's theorem), establishes that for any given degree of noise in a communication channel, it is still possible to communicate (digital) data nearly error-free up to a computable maximum rate by introducing redundancy into the channel. This also suggests why nature has chosen a clearly digital code for life in the form of nucleic acids (e.g. DNA and RNA). On the other hand, while the balance was positively reached, that is, the accuracy in the replication process was greater than the incidence of error, some errors did occur, and these could also be quantified using classical information theory.

While the nature of the variations (of which some can be identified as errors) can be attributed to noise, the correction is nothing but a mathematical consequence. If errors prevail, both in the primary source and replicants, then the cells and organisms have a greater chance of dying from these variations. On the other hand, new variations may confer survival or reproductive advantages in changing environmental conditions and, thus, will be positively selected during evolution. Natural selection hence effectively reprograms nature, cells and organisms. Of course this does not explain how the whole process began, the origins of life being still an open question. 

Approached as computer programs, one can explain how certain patterns, e.g. the information content in molecules, such as DNA or RNA, may have been produced in the first place~\cite{zenilfqxi,zenilalgo,zenilnaturalcomputing,zenillife,zenilentropy}. There is also the problem of the uncaused first computer program, given that the process generating information and eventually computation required computation in the first place (e.g. the very first laws in our universe). In fact, we have explored these ideas, taking seriously the possibility that nature computes, and we have suggested a measure~\cite{zenilpt} of the property of being ``computer-" or ``algorithm-like", the property of \textit{algorithmicity}. In~\cite{zenilalgo} we undertook a search for statistical evidence, with interesting results.

Hacking strategies have been in place in biological systems since the beginning, viruses for example, are unable to replicate by themselves, but they trespass the cell membrane and release their DNA content into the cell nucleus to be replicated. Cells are the basic units of life because of this property, they are the smallest unit that fully replicates by itself. No simpler unit is able to replicate in full like the cell even if RNA is suspected to be a possible precursor of replication, and possibly the first self-replicating molecule because RNA can both store information like DNA but also fold similarly to proteins and therefore serve as building blocks to build more complex structures.

Viruses are clearly (in consensus among biologists) non-living structures of encapsulated DNA that evolve by undergoing a process of natural selection and are evidence of non-living matter  subject to the same process. Though they may appear so, viruses have no self-purpose or will of their own, neither as individuals or as a group. Indeed, evolution by natural selection is a mathematical inevitability independent of substrate. Despite the common flawed fashion in which the concept of evolution by natural selection may popularly be conveyed --- as a biological process that carries a self-purpose meaning as an objetive function for self-preservation --- it is a simple statistical consequence that cannot be contested. In the case of viruses, for example, it is the DNA in the virus that make it more effectively into the nucleus and replicates faster that will dominate. This basic consequence is the basis of the mechanism of evolution by natural selection. By adding limited resources, such as the number of cells that can be infected, one can extrapolate this simple mathematical fact to anything else other than viruses. Even if viruses are naturally perfect carriers of computing code, biological programming and reprogramming is in general very similar because we know everything is encoded in the cell's genetic material.

\subsection{A hacker view of cancer and the immune system}

Equipped by these mechanisms of evolution by natural selection and the way in which self-replicating cells can be hacked a hack to fight diseases has been devised in the efforts to treat cancer. Traditional drug and radiation therapies have not been very successful so far but so-called \textit{gene therapy} is promising spectacular results. \textit{Gene therapy}  is a mechanism to deliver specific genes into tumors by using similar, if not exactly, the same strategies to that of viruses in their delivery of their payload. Indeed, the approach relies heavily on using viruses to deliver anti-tumor genes into the target cancer cells. Various hacking approaches using viruses can be very devised. If the foreign code enters the nucleus it will be replicated, but make a buggy virus that only reaches the cytoplasm after the cell membrane and it can interfere with protein transcription (making the enzymes in the cytoplasm to believe that the instructions come from either legitimate external signals from other cells or from the regular instructions coming from the cell nucleus). The fake instructions will then modify the cell behaviour (function) with the nucleus remaining intact and replication of its original untouched code guaranteed. Indeed, one can see how genes are effectively subroutines, computer programs that regulate the type (shape) and number of building blocks (proteins) to be produced that in turn fully determine the cell function(s). Let the instructions reach only outside the nucleus and they will interfere with messages (RNA) only indicating protein production, or do so inside the nucleus and modify the actual genome of the cell (DNA) to replicate it (what viruses do, integrating their DNA content into the host genome and undergoing a coevolution~\cite{virolution}).

There is therefore a potential to develop these ideas into a more systematic information-theoretic and software-engineering view of life, cancer and immune-related diseases based on these amazing purely computational mechanisms in biology. While it is clear that information processing is in a very fundamental sense key to essential aspects of the biology of different species (see Fig.~\ref{programmability} top), by applying ideas related to computability and algorithmic information theory we can take a step ahead to view processes from fresh and different angles to conceive new strategies to modify the behaviour of cells against diseases.

Cancer, for example, is like a computer programming bug that does not serve the purpose of the multicellular organism. Cancer cells are ultimately cells that grow uncontrollably and do not fulfill their contract to die or stop proliferating at the rate at which they must if the multicellular host is to remain stable and healthy. From a software-engineering perspective it is a cellular computer program that has gone wrong, resulting in an infinite loop with no halting condition, hence in effect being out of control. Normal cells have a programmed mechanism to stop their cell cycle at the point at which the population of previous cells is replaced. When the cell cycle produces more than the number of cells necessary to replace the population, the result is a mass of abnormal cells that we call a tumour. 

Cancer can be seen as a purely information-theoretic problem: the information dictating the way in which a cell replicates is compromised, either because it, as it were, reneges on a contract with the multicellular organism, resulting in the cell behaving selfishly and replicating with no controls, or else because noise in the environment in the form of external stimuli has broken the programming code that makes a cell's behaviour ``normal". In other words, it is either a bug or a broken message. The question is therefore how a cell can transfer its information at replication time without generating instructions that produce cancer states when the said cell encounters noise. 

Likewise, the immune system can be seen as an error-correcting code. One key aspect of the immune system is diversity which is largely contributed by T and B lymphocytes, cell types of the adaptive immune system. By gene rearrangement of segments in their antigen receptor genes, a highly complex repertoire of different receptors expressed by individual B or T cell clones is generated, which are specific for different antigens. The B cell antigen receptor, surface bound or secreted as an antibody molecule by terminally differentiated B cells, recognises whole pathogens without any need for antigen processing, while the T cell antigen receptor is exclusively on the T cell surface and recognises processed antigens presented on so-called MHC molecules.

When B cells and T cells are activated and begin to replicate, some of their progeny become long-lived memory cells, remembering each specific pathogen encountered, and can mount a strong, faster response if the pathogen is detected again. This means the memory can take the form of either passive short-term memory or active long-term memory.

The memory of the immune system stores all the information relating to all the pathogens we have encountered in our lives~\cite{memory}. When B cells and T cells are activated some will become memory cells. Throughout the lifetime of an animal these memory cells effectively form a ``database" of effective B and T lymphocytes~\cite{memory2}. Upon interaction with a previously encountered antigen, the appropriate memory cells are selected and activated. 

The major functions of the acquired immune system that involve information include:

\begin{itemize}
\item Pattern recognition:``non-self" antigens in the presence of ``self", during the process of antigen presentation.
\item Communication: Generation of responses that are tailored to maximally eliminate specific pathogens or pathogen-infected cells.
\item Storage (memory): Development of immunological memory, in which\\pathogens are ``remembered" through memory cells.
\end{itemize}

Yet another illustration of how the immune system can be seen from a purely informational perspective is the fact that newborn infants have no immune memory, as they have not been exposed to microbes, and are particularly vulnerable to infection. Several layers of passive protection are provided by the mother. During pregnancy, antibodies of the IgG isotype are transported from mother to baby directly across the placenta~\cite{sedmak}, so human babies have high levels of antibodies even at birth, with the same range of antigen specificities as their mother. Protective passive immunity can also be transferred artificially from one individual to another~\cite{transfer}. 

This brief account reveals the extent of the role played by information in the immune system. Immune-related diseases are related to informational dysfunction, such as wrong signaling, defects in immune tolerance or misguided pattern recognition. This can cause the immune system to fail to properly distinguish between self and non-self and attack part of the body, leading to autoimmune diseases. As described before, an important role of the immune system is also the identification and elimination of tumours. The transformed cells of tumours, often express antigens that are not found in normal cells. The main response of the immune system to tumours is to destroy the abnormal cells. Tumour antigens are presented on MHC class I molecules in a similar way to viral antigens, and this allows killer T cells to recognise the tumour cell as abnormal and kill it. The immune system produces and reprograms cells to match them with pathogens. Antigens are messages. A vaccine is a way to send a fake message without the actual content.

A model that a team created~\cite{fedishev} shows that the stability of a gene network stems from several major factors. These factors include ``effective'' genome size, proteome turnover, and DNA repair rate but also gene network connectivity. The researchers concluded that hacking any of these parameters one could increase an organism lifespan, an hypothesis supported by the biological evidence~\cite{fedishev}.

Cellular death has a strong information-theoretic component, evident in the way evolution has programmed multicellular organisms. Cancer, and indeed laboratory cell lines (the cells used in labs) are immortal~\cite{skloot}; they continue dividing indefinitely (just as stem and germ cells too). An immortal living cell can only last as long as its error-correcting methods (such as conserving the length of its telomeres, up to the so-called \textit{Hayflick limit}~\cite{hayflick1,hayflick2}, and correcting DNA mutation mechanisms) preserve genome integrity and stability. In practice this is impossible, for purely chemical and physical reasons. For instance, DNA is believed to undergo more than 60\,000 single mutations a day  in the genomes of mammalian cells~\cite{mutations,mutations2, mutations3}, and a fraction of DNA will never be repaired by specific DNA enzyme-repairing molecules, hence leading to genome instability and degenerative diseases such as cancer~\cite{corcos}, and neuronal and neuromuscular diseases~\cite{subba}. Telomeres, for their part, cannot remain unchanged due to chemical degradation, so after about 40 to 60 divisions, they shorten and die~\cite{hayflick1,hayflick2}. But all cells are potentially immortal. If they die at the end of a generation, it is because they are programmed to do so in order to serve as building blocks for multicellular organisms. The length of a cell's telomeres is also believed to play a fundamental role in the fight against cancer: the length of the telomeres may be such that they are shorter and degrade faster than the critical number of accumulated mutations, hence effectively preventing a cell from continuing to divide and replicate its errors before dying~\cite{eisenberg,shay}. We can thus start seeing how information theory and effective programming are implicated in the most basic mechanisms of life and death. Hacking cells can be achieved simply by inhalation just as viruses do, and it is promising to be effective in a trial with Cystic fibrosis caused by a gene mutation located in the chromosome 7. The hack consists in introducing the normal gene into fat globules (liposomes) that deliver the gene into the cells in the lung lining~\cite{alton}.

\subsection{Immunity as computation and cancer as a software-engineering problem}
\label{classificationcancer}

\begin{figure}
\centering
\scalebox{.33}{\includegraphics{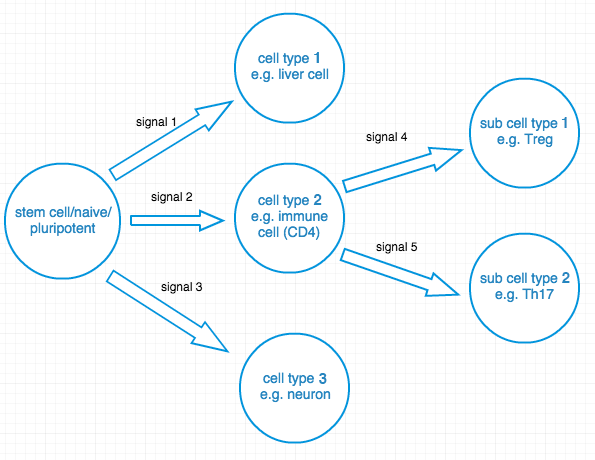}} \scalebox{.3}{\includegraphics{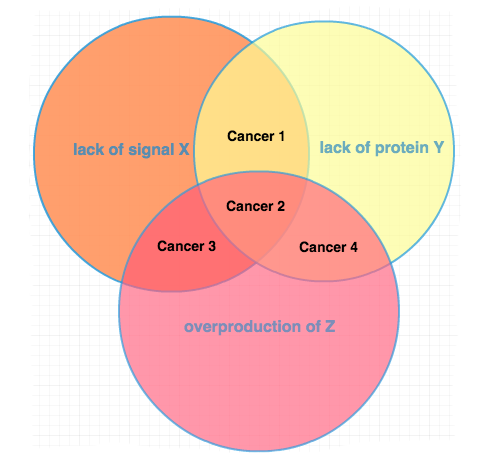}}
\caption{\label{cellprogramming}Left: Cells are like computer programs ready to be reprogrammed, by nature or artificially, to become other cells. Cells usually start from a cell that is amenable to being programmed, as it has no specific function other than precisely to become a new type of cell and to survive until doing so by receiving the correct signals from its environment (including other cells). In this cartoon, signals 2 and 5 can produce, for example, a Th17 (T helper 17) starting from a pluripotent cell, a cell that can potentially become many different types of cell. The pseudocode to reprogram a cell involves calling computer subroutines (chemical signals) 2 and 5. End points in the tree can be seen as computer programs reaching a stable configuration, an attractor in a dynamical system that can be either a fixed point, a cycle or strange attractor that cannot easily be reprogrammed back to previous developmental stages. Right: A computational reclassification of cancer and tumour profiling. Intersecting Venn diagrams of bug state spaces identified as leading to different diseases and tumour types (e.g. lack of signal X, lack of protein Y, overproduction of Z). This would immediately suggest the possibility of more tumour types (cancers) than source bugs (there are 4 intersections from 3 potential interacting source bugs), but the important message here is that these bugs can be debugged by type, and tackling bug 1 can actually fix many types of cancers if they are only triggered by the presence of all these bugs. This enables more general strategies than attacking each cancer in isolation, as an instance of a  type (e.g. by tissue, as has been the practice to date) or by using extreme approaches such as radio- or chemotherapies that are anything but localised or personalised.}
\end{figure}

Cells are continually programmed and reprogrammed by the environment and by the cells themselves. Fig.~\ref{cellprogramming} (left) illustrates how this process happens. In the example, generic signal 2 first turns a stem cell into an immune system cell (e.g. CD4-positive T helper cell), and then signal 4 finally drives it to a ``steady state", where it has completely differentiated into a cell with a specific function. This is an oversimplification of a complex system, as information and signals do not flow in a single direction. In fact T helper cells in the immune system regulate other immune cells and thus send back signals to the reprogramming pipeline according to the signals they themselves receive from other cells in the body and depending on the environment (e.g. pathogens present). This also implies that one can reprogram cells with other cells, and that differentiated cells, though more difficult, are not generally impossible to reprogram. In fact, plasticity of CD4 T cell subsets has been observed in several directions~\cite{plasticity}. Signals 2 and 4 may also prompt other cells to react in different ways. Signals are simply chemicals or chemical cocktails. For example, programming a T cell towards the regulatory T cell (Treg) subset requires proteins such as the transcription factor Foxp3 (forkhead box P3)~\cite{foxp3} and the cytokines TGF-$\beta$~\cite{tgf} (transforming growth factor beta) and interleukin-2 (IL-2) that the body itself produces using the same or other cells, in effect producing its own signals for reprogramming itself. However, the same signals can also be artificially produced. A model based on cellular automata has shed light on the way in which the immune system can be interpreted as a computer program subject to sensitivity and robustness in the face of external changes (signals, mutations), quantifying the number of error that keep or remove the cellular automaton from its regular basin of attraction and therefore changing or maintaining its original function interpreted as the steady state of the cellular automaton seen as a dynamical system~\cite{cs}.

As a programming strategy, one needs for example to induce the production of Foxp3 and TGF-$\beta$ in a naive CD4 T cell to produce a Treg-like cell (under ideal conditions and in the presence of IL-2). Importantly, immune system signals also contribute to the regulation of cancer development. In fact, it is believed that TGF-$\beta$ is related to cancer. Acting through its signaling pathway, it can stop the cell cycle in order to prevent proliferation, induce differentiation and promote apoptosis~\cite{cancer2}. 

What is known as \emph{morphological computation} is based on the universal phenomenon in biology that form is function, that is, the shape of biological structures (epitomized by proteins) determine the structure's biological function. While all forms of computation can be traced all the way back to classical computation and the concept of computational universality, which blurs any essential boundary between software and hardware, biology is the quintessential example. Indeed natural and morphological/spatial computation is the realisation that in physics but mainly in biology hardware is also software. Everything is in some way a kind of embedded computation. For example, DNA is code and storage but other molecules are not only code and storage devices but also functional devices, such as RNA and proteins, these latter  carrying different structural information with it. Not only does the code depend on the nucleotides but, in the DNA configuration and packaging, on histones and other structural proteins that enable or hinder fragments of code from replicating; the structure of DNA itself encodes information. RNA, in turn, is a message when encoding for proteins and an output when acting as regulator, and proteins can regulate translation and gene expression.

Using a combination of the fundamentals of information theory and the principles of natural selection one may grasp how the immune system seems to have come into existence, it being a natural error-correcting mechanism continually streamlined and reprogrammed by natural selection. This latter is again constrained as to what  is possible according to the Shannon limit, determined by Shannon's aforementioned Noisy-channel coding theorem~\cite{cover}, for if the replication process exceeds this limit, then its accuracy will be compromised and there will be a loss of  information, eventually leading to failure to recode basic functions of living organisms such as genes. 

How does nature reprogram (or \textit{un-program}, from the point of view of their original function) cells so that they become cancerous?  There are various hypotheses, one of which posits the breaking of a ``contract" that cells made when they went from being unicellular life forms, hence "selfish", to being components of multicellular organisms such as the human body. The existence of tumour cells that lose certain functions but remain able to replicate is an indication that life is highly hierarchical and modular and robust, unlike traditional artificial computer programs that are very vulnerable to random errors introduced into their source code. In a sense, when it comes to living organisms we seem to be dealing with a streamlined version of an object-oriented programming language.

Methods for modifying the functioning of cells have already shown promise. Recently, for example, it has been shown that the immune system can be reprogrammed to fight cancer~\cite{lizee}. As an example, in~\cite{reviewcancer}, it is shown that genetically modified T cells can be used in cancer therapy. Indeed, as is pointed out, tumour cells (tumours) follow many strategies to evade the immune system, including tampering with genes that would normally regulate a function related to the sensitivity of an immune cell to certain signals produced by said tumour cells~\cite{schreiber}. However, tumour cells generate an immunosuppressive tumour environment that leads the immune system to neglect the tumour and therefore the great danger to the host multicellular organism (the host body)~\cite{reviewcancer}. 

So how might we reprogram tumour cells themselves or immune cells to recognise and naturally fight tumour cells? For a cell to replicate it needs to become redundant in the face of environmental noise, hence more simple. We can target the genes that in the replication process contribute less to the information content of the cell because of its state of redundancy. In other words, tumour cells may be identified because they are less sensitive and therefore also less programmable (just like the examples at the top and middle of Fig.~\ref{fig3}). 

The current common classification of cancer divides it into broad groups that are not related to the type of possible error leading to a cancer but rather to where a cancer originates and other physical properties. However, it has been found that different types of cancer under this classification can be deeply related, possessing similarities that cut across different types of tissue ~\cite{heim}. This may be because certain cancers may have a set of common causes related to a type of bogus program. In software terms, such a classification scheme amounts to classifying all operating system errors in the same bin simply because they involve operating systems, and not because, say, one has a software bug related to reference software bugs (memory access violations, null pointer dereference), another an arithmetic error (e.g. division by zero, precision loss or overflow), and still others have logic software bugs (e.g. infinite loop recursion) or syntax software bugs. 

\begin{figure}
\centering
\scalebox{.27}{\includegraphics{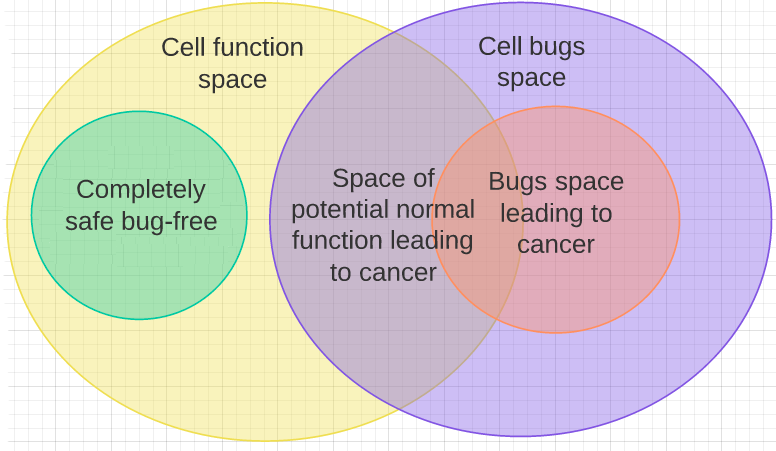}}\\
\bigskip
\bigskip
\medskip

\scalebox{.2}{\includegraphics{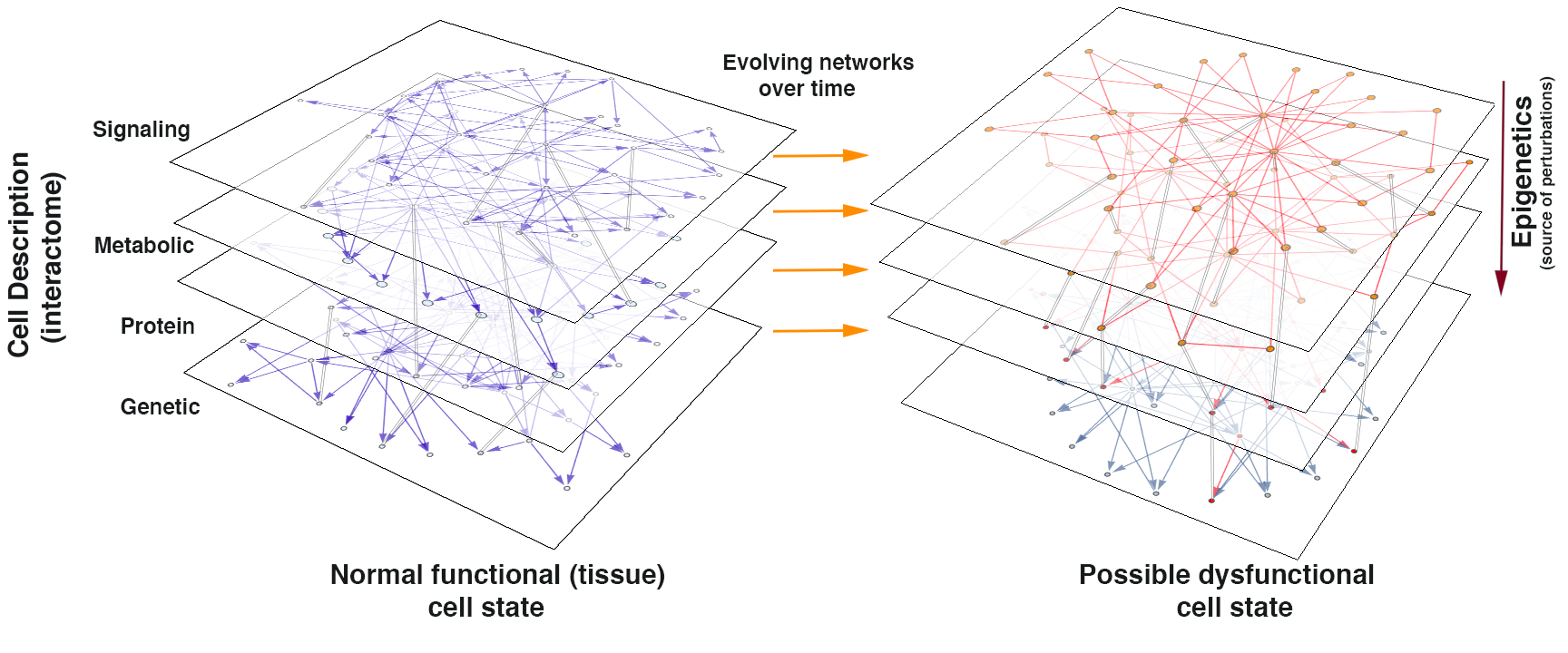}}
\caption{\label{networks}Top: Normal function (yellow) vs. bug space (blue) or abnormal cell function. First questions about cell biology, disease and cancer from the software-engineering perspective is how large each of these sets really is relative to each other. Natural selection provides some hints, the bug space cannot be larger than the normal function space before and until reproductive age when we are subject to evolutionary pressure (natural selection simply says: if the bug space is larger than the healthy space then we would be all dead because population would shrink at every generation). Directions in the networks represent a gene (transcription factor) regulating other genes e.g.  upregulating, i.e. increasing its production, or downregulating, i.e. decreasing its production; or proteins interacting or metabolites reacting or signals being transported from one element to another. Bottom: Propagation of undesired code in a network of networks: Several cell networks at different levels of description form another causal network. The full description is also called the cell \textit{interactome}. Changes to nodes (genes, proteins, metabolites, signals) or edges (regulation, reaction, channel) in any layer lead to functional and description changes, and some of these will lead to a dysfunctional (diseased) cell, such as a tumour cell, where a normal tissue cell behaves differently than expected (e.g. uncontrolled replication). All these changes are reflected in the \textit{interactome} and therefore in the description, and in the information content, of the cell.}
\end{figure}

Some of these software bugs may lead to incremental error accumulation. We think these types of software errors may prompt the rethinking of current cancer classifications, leading to the grouping of cancers in terms of their information/computational type, i.e. the type of error that leads to cell reprogramming, and not by their point of origin. Such an approach would make it possible to explain why so many cancers have so many things in common, while cancers that are supposed to be of the same type, e.g. liver cancers, can actually be very dissimilar. A classification strategy based on place of origin and not type of error is likely to fail because it fails at characterising the cause of the problem. It may be compared to trying to debug a program based on its application, for instance, treating all software bugs in word processors similarly, or all bugs in animation software using a common strategy, which makes no sense. It is starting to be more widely recognised that cancer should not be classified or studied by tissue type (which immediately leads to thousands of cancer types, if one takes cell subtypes into consideration), but by bug type. Not any bug will turn a cell into a cancer cell, and we just do not know how many will. This is perhaps one of the first quantification tasks and one of the first interesting research questions --- to determine the bug space (see Fig.~\ref{cellprogramming} right and~\ref{networks} top) leading to different diseases and tumour types. We have suggested that this can be more systematically approached by following the behavioural black box approach that we have illustrated. Not because we have a penchant for seeing things as black boxes, but because as we have explained, science has to deal with black boxes, given certain limitations that go beyond the pragmatic and are fundamental and intrinsic to even the most simple deterministic rule-based systems. Fig.~\ref{cellprogramming} shows a Venn diagram depicting the proposed informational view of a software-engineering classification of diseases such as cancer, based on bug type rather than tissue origin. The diagram is in itself a simplification. For example, in biology a lack of signal X likely represents a lack of production of a protein or its overproduction, hence bugs may themselves be produced by other bugs. But here they are considered only as direct causes, which one has to tackle layer by layer just as one would in a traditional computer program, inserting debugging breakpoints either to identify the primary cause or in order to design a patch  ---  just like a software patch  ---  to stop the propagation of the error before the first forking path leading to the undesired behaviour. Reprogramming from scratch is of course always more desirable than a patch, which can introduce new undesirable effects. Indeed this can be the informatics definition of a drug with its common secondary effects. Hence we would wish to move from traditional drugs to drugs closer to the causes.

Some  interesting initial software-engineering questions about cell biology, bugs and cancer can be found illustrated in Fig.~\ref{networks} (top). While the potential cell bug space can be very large (blue), there have to be natural selection mechanisms that prevent the bugs leading to fatal diseases (e.g. red set), keeping them small, at least before and during reproductive age. Later in life, because there is no longer evolutionary pressure, the size of these spaces can increase freely. For example, experience with cancer tells us that about 1/3 of the population (believed to rise to 1/2) can be considered likely to reach some cancer state during their lifetime. The immune system as well as cell-intrinsic mechanisms prevents normal cells from proliferating when they reach the bug space, yet cells and cell populations are constantly, dynamically moving along these sets. What we have discussed here suggests that one cannot find infallible methods to keep cells in a completely safe bug-free zone (green). More complicated questions follow, questions about whether the normal functioning of the cell can dynamically lead to cancer (large and small intersections) even without the introduction of mutations. That is, whether cells may be programmed to eventually reach some sort of tumour state, even if they do not do so for a long initial period of time (e.g. before reproductive age). 

\begin{figure}
\centering
\scalebox{.43}{\includegraphics{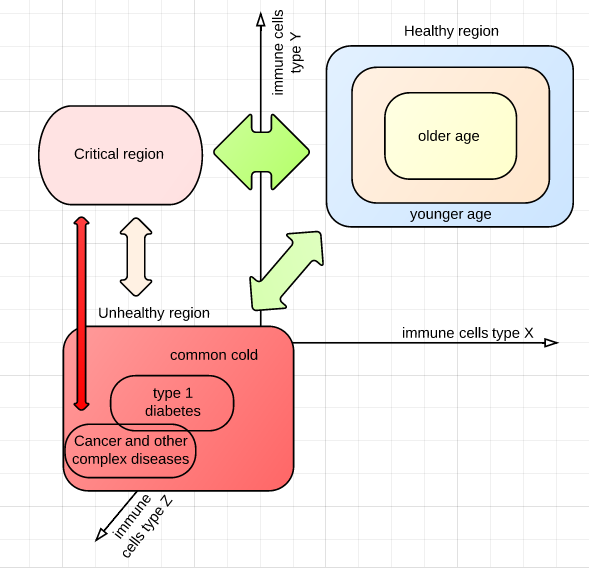}}\\
\caption{\label{phasespace}High dimensional phase space of the immune dynamical system: The main characteristic of the space is the existence of stable regions that represent the healthy or unhealthy state as fixed or strange attractors of the dynamical system of the immune system that is perturbed over time (both by weakening self-defenses or by introduction of foreign organisms, such as virus, bacteria and dysfunctional cells such as tumours). Natural selection establishes that the healthy region in early ages is necessarily greater than the unhealthy region that leads to death hitting the reproductive rate (e.g. the cancer region). However, after reproductive age there is no evolutionary pressure and the unhealthy region grows over time. For example, cancer is a more common disease in elderly people, which is however also related to the fact that usually several mutations have to be acquired over a relatively long time period in order to cause cancer. The cell state of the body crosses a critical region many times during the lifetime of an organism and whether the organism goes in one or another direction depends on the strength and capability of the host immune system. External factors perturbing the state of an organism over time lead to changes in this high dimensional space of immune cell types that usually respond to different threats in different ways, both in number of cells and their type and  are hence excellent candidates for disease monitoring. With some probability the organism leans towards the unhealthy region or back to the healthy one, depending on the strength of the immune system and the degree of invasion. A subset of the unhealthy region is due to complex diseases from which it is difficult to get back to the critical region towards other equilibria (e.g. healthy state) but not impossible.}
\end{figure}

Diseases imply deviations of a cell towards pathological states that are encoded in the cell's descriptions. Fig.~\ref{networks} (bottom) shows the intricate ways in which different descriptions of the cell interact with each other. All these causal interactions fully describing the cell constitute what is called the \textit{interactome}. The interactome and each of its complicated interacting parts can be studied with tools from classical and \textit{algorithmic information theory} (AIT). AIT is the subfield that characterises lossless compression, and has as its core Kolmogorov complexity. It is the bridge between computation and information and it deals with an objective and absolute notion of information in an individual object, such as a cell in a network representation~\cite{zenilbibm}. We have proposed ways to study and quantify the information content of biological networks based on the related concept of algorithmic probability, and we have found that it is possible to characterise and profile these networks in different ways and with considerable degrees of accuracy~\cite{zenilgraph,zenilkiani}. This shows that the information approach may open a new pathway towards understanding key aspects of the inner workings of molecular biology  causality-driven rather than correlation-driven.

Important sources of information are epigenetic phenomena, an additional layer of complexity reversing the traditional molecular biology dogma that describes how information is transferred from the genome all the way to the upper levels. Epigenetics shows that information can flow bottom-down from all upper layers to the lower layer (genome). This information coming back to the cell alters how genes are finally expressed even if the cell's genome (the DNA code) remains exactly the same. Epigenetics is the process in which external information is introduced to the various cell layers that can disrupt the function of it and lead to changes including diseases. One can study the rate of information propagation by looking at the different layers and then identify the layer source of the disfunction. Some changes will propagate fast inducing a different dynamic with attractors that are different to those of the normal cell function, other changes may not propagate thanks to the qualitative robustness of biological systems (see Fig~\ref{conformationalspace} (top)). Small changes in lower layers are amplified in upper layers, e.g. SNPs (single nucleotide polymorphism)~\footnote{A DNA sequence variation occurring commonly within a population (e.g. 1\%) in which a single nucleotide A, T, C or G in the genome (or other shared sequence) differs between members of a biological species or paired chromosomes.} in the gene BRCA1 can cause a cell to become cancerous due to defects in DNA repair, leading to additional mutations in the cell that ultimately can cause its transformation to a cancer cell. This is why complex diseases may be better detected at the level of the upper layers while "simpler" diseases such as cancer induced by cancer-predisposing BRCA1 SNPs (a very specific case) can be detected at the lowest level. The network on which these changes propagate is poorly understood as it requires integration of different data sources put together and studied over time, which is the direction in which we are heading to by studying the \textit{interactome} (the cells full description in the form of a network of networks) over time, the ultimate goal of \textit{network reconstruction}.

\section{Conclusions}

We have seen that uncomputability prescribes limits to what can be known about nature or models of nature, limits that are likely to apply to natural and biological systems or the models we build of them, and therefore we cannot help but develop an encompassing behavioural approach that can ultilise ideas and tools from both theoretical computer science and software engineering. These ideas can then be further developed to yield new concepts, classifications and tools with which to reprogram cells against diseases. This is a direction in which biology is moving by adopting new immunotherapies against diseases such as cancer.

 An information computational approach to cancer and human diseases may be key to understanding molecular medicine from a new perspective. The most promising approach, as in software-engineering at the design stage, may involve prevention through permanent monitoring of the immune system (see Fig.~\ref{phasespace}) based on systematic screening of the immune status over time in order to detect trends in responses to disease, including early tumour detection. It is now clear that the best way to detect early signs of disease must inevitably involve tracking the unfolding computation of the immune system in the course of its normal operation. If the immune system produces more cells of a certain type, it is a sign of change and likely a reaction to a threat, and the trajectory of immune system behaviour is an indicator of the direction an individual's health will evolve in. The immune system as a computational device, being our potentially best debugging system, is key to our strategy for defeating cancer and many other diseases, and thereby reprogramming our fate. Of course debugging software can break down too, just as may happen in an artificial software debugger, and this is where autoimmune and inflammatory diseases come into play. These can likewise be detected by looking closely at the behaviour of the immune system.

\section*{Acknowledgements}

We wish to thank the rest of the Unit of Computational Medicine team at Karolinska Institutet, and the support of AstraZeneca, the Strategic Area Neuroscience (StratNeuro), the Foundational Questions Institute (FQXi), the John Templeton Foundation and the Algorithmic Nature Group, LABORES.

\newpage

\textsc{Hector Zenil} (BSc Math, UNAM; Masters Logic, Paris 1 Sorbonne; PhD Computer Science, Lille 1) has held positions at the Behavioural and Evolutionary Lab, Department of Computer Science, University of Sheffield and at the Structural Biology Group at the Department of Computer Science, University of Oxford in the UK; and at the Unit of Computational Medicine, Science for Life Laboratory (SciLifeLab), Centre of Molecular Medicine, Karolinska Institute in Stockholm, Sweden. He is also the Head of the Algorithmic Nature Group and member of the board of directors of LABoRES in Paris, France. He has been visiting graduate student at the Massachusetts Institute for Technology (MIT), invited visiting scholar at Carnegie Mellon University (CMU), and invited visiting professor at the National University of Singapore (NUS). He has also been a senior research associate and external consultant for Wolfram Research, an invited member of the Foundational Questions Institute (FQXi), and a member of the National Researchers System (SNI) of Mexico.\\

\bigskip

\textsc{Angelika Schmidt} studied Biology at the Technical University of Darmstadt, Germany and was a visiting scholar at The Rockefeller University, New York, USA. She obtained her PhD in Immunology from the University of Heidelberg, Germany on a project about regulatory T cells in Peter Krammer's group at the German Cancer Research Center (DKFZ). She is currently Postdoc and Marie Curie Fellow in the Unit of Computational Medicine, Science For Life Laboratory (SciLifeLab) and the Centre for Molecular Medicine at Karolinska Institute in Stockholm, Sweden, focusing on regulatory T cell induction and suppressive function. \\

\bigskip

\textsc{Jesper Tegn\'er} is Chaired Strategic Professor in Computational Medicine at the Centre for Molecular Medicine and Sciences For Life Laboratory (SciLifeLab) at Karolinska Institutet and Karolinska University Hospital in Stockholm, Sweden. He heads a research group of 35 people, one-third working in the molecular biology lab and two thirds working as computational experts. He was a Visiting Scientist and Postdoctoral Fellow (Harvard) on a Wennergren five-year research position, 1998-2001 with an Alfred P. Sloan Fellowship for research. He has authored over 100 papers, including technical computational papers as well as medical publications (Science 2005, Nature Genetics 2009, PNAS 2002, 2003, 2004, 2007 and 2009, Cell 2010). He holds three separate undergraduate degrees (MedSchool, MSc, Mathematics, MSc, Philosophy), Ph.D./M.D. 1997 Medicine. 

\end{document}